\documentclass{article}

\usepackage[preprint,nonatbib]{neurips_2020}
\usepackage[utf8]{inputenc} 
\usepackage[T1]{fontenc}    
\usepackage{hyperref}       
\usepackage{url}            
\usepackage{booktabs}       
\usepackage{amsfonts}       
\usepackage{nicefrac}       
\usepackage{microtype}      
\usepackage[pdftex]{graphicx}
\usepackage{amsmath,amssymb} 
\usepackage[ruled,vlined, linesnumbered]{algorithm2e}
\usepackage{import}
\usepackage{mathabx}
\usepackage{color}
\usepackage{bbold}
\usepackage{breqn}
\usepackage{svg}
\usepackage{multicol}
\usepackage{multirow}
\usepackage{enumitem}
\usepackage{tikz}
\usetikzlibrary{spy}
\usetikzlibrary{arrows,shapes.geometric,positioning}
\usepackage{capt-of}
\usepackage{subfiles}
\usepackage{amsmath}
\usepackage{mathtools}
\usepackage{comment}
\usepackage{wrapfig}
\usepackage{caption}
\usepackage{subfig}
\usepackage{adjustbox}
\usepackage[demo,abs]{overpic}
\usepackage{titlesec}

\usepackage{animate}

\definecolor{darkgreen}{RGB}{30,150,30}
\definecolor{darkblue}{RGB}{0,0,127}
\definecolor{darkyellow}{RGB}{171,133,0}
\definecolor{darkred}{RGB}{180,20,20}
\definecolor{darkmagenta}{RGB}{127,0,127}
\definecolor{darkcyan}{RGB}{0,127,127}

\newif\ifdrafting 
\draftingfalse 

\ifdrafting
  \newcommand{\OG} [1] {\textcolor{darkgreen}{[OG: #1]}}
  \newcommand{\VJ} [1] {\textcolor{darkblue}{[VJ: #1]}}
  \newcommand{\TA} [1] {\textcolor{darkmagenta}{[TA: #1]}}
  \newcommand{\TODO} [1] {{\color{darkcyan}{\bf [TODO: #1]}}}
   \newcommand{\checkthis}[1]{#1}
\else
  \newcommand{\OG} [1] {}
  \newcommand{\VJ} [1] {}
  \newcommand{\TA} [1] {}
  \newcommand{\TODO} [1] {}
  \newcommand{\checkthis}[1]{#1}
\fi

\newcommand{\s}{$\mathcal{S}$}
\newcommand{\mps}{$\hat{\mathcal{S}}$}
\newcommand{\mpa}{$\hat{\mathcal{A}}$}

\newcommand{\ls}{$\tilde{\mathcal{S}}$}
\newcommand{\la}{$\tilde{\mathcal{A}}$}

\graphicspath{{images/}{../images/}}

\title{Generative View Synthesis: From Single-view Semantics to Novel-view Images}

\author{%
Tewodros Habtegebrial$^{1,4}$ \quad Varun Jampani$^{2}$ \quad Orazio Gallo$^{3}$ \quad Didier Stricker$^{1,4}$ \\
$^1$\texttt{TU Kaiserslautern} \quad $^2$\texttt{Google Research} \quad $^3$\texttt{NVIDIA} \quad $^4$\texttt{DFKI}
}

\makeatletter
\@addtoreset{section}{part}
\makeatother
\titleformat{\part}[display]
{\normalfont\LARGE\bfseries\centering}{}{0pt}{}
\begin{document}
\maketitle

\begin{abstract}

Content creation, central to applications such as virtual reality, can be a tedious and time-consuming.
Recent image synthesis methods simplify this task by offering tools to generate new views from as little as a single input image, or by converting a semantic map into a photorealistic image. 
We propose to push the envelope further, and introduce \emph{Generative View Synthesis} (GVS), which can synthesize multiple photorealistic views of a scene given a single semantic map.
We show that the sequential application of existing techniques, e.g., semantics-to-image translation followed by monocular view synthesis, fail at capturing the scene's structure.
In contrast, we solve the semantics-to-image translation in concert with the estimation of the 3D layout of the scene, thus producing geometrically consistent novel views that preserve semantic structures.
We first lift the input 2D semantic map onto a 3D layered representation of the scene in feature space, thereby preserving the semantic labels of 3D geometric structures. 
We then project the layered features onto the target views to generate the final novel-view images.
We verify the strengths of our method and compare it with several advanced baselines on three different datasets.
Our approach also allows for style manipulation and image editing operations, such as the addition or removal of objects, with simple manipulations of the input style images and semantic maps respectively. \\
Visit the project page at \url{https://gvsnet.github.io/}.

\end{abstract}


\section{Introduction}
\label{sec:intro}
\vspace{-2mm}

The rising demand for digital content, together with the widespread availability of high-quality digital cameras, has fueled the need for tools and algorithms to democratize content creation.
A prominent example of one such technology is novel view synthesis, which allows the artist to render a scene from new viewpoints using as few as two images~\cite{choi2019extreme, zhou2018stereo}, or even just one~\cite{wiles2019synsin}.
Photorealistic images can also be generated by editing a simplified representation of the scene, such as a semantic map, followed by image-to-image translation~\cite{spade}, but the viewpoint cannot be manipulated.

In this work, we propose \emph{Generative View Synthesis} (GVS), which combines the advantages of both approaches.
Given a single semantic map, which is easy to edit and requires no image capture, GVS can generate RGB images of the same layout, but from new, arbitrary viewpoints.
Not surprisingly, GVS also inherits the challenges of both: generating RGB values from a bare semantic map is an ill-posed problem that is further complicated by the need for the different output views to be photometrically and geometrically consistent. 
One could tackle this problem with a sequential application of existing techniques. That is, we can first convert the single-view semantic map into an RGB image using image-to-image translation techniques~\cite{pix2pix}, and generate novel RGB views using monocular novel view synthesis techniques~\cite{wiles2019synsin}.
However, we observe that this may fail at preserving the scene's structure accurately, as shown in the animation in Figure~\ref{fig:teaser}.

\begin{figure}[ht!]
    \centering
    \animategraphics[width=\textwidth]{2}{images/teaser/teaser_mirror_}{00}{07}
    \caption{\textbf{Generative View Synthesis} is a method to generate photorealistic images from novel viewpoints, given just a semantic map and a style image. Here we show lateral ($l_{0-7}$) and forward ($f_{0-7}$) camera motion. Because no methods exist to solve this problem, we propose to use SPADE~\cite{spade} followed by single-image MPI rendering~\cite{zhou2018stereo} as a baseline. Our method better preserves thin structures and produces geometrically consistent views. \textit{Animated figure. Please view in Adobe Reader and click on the image to see the animation. Other PDF viewers may have issues, in which case please refer to the supplementary.}}
    \label{fig:teaser}
\end{figure}

Our key insight is that semantic maps are particularly informative about the structure of a scene, despite offering no information about its photometric properties.
Semantic segments, in fact, carry explicit and unambiguous information about occlusion boundaries.
This is in stark contrast with RGB images, where edges can also result from texture.
We leverage this observation to preserve geometric consistency between multiple output views. 
Specifically, instead of converting the semantic map to an RGB image, we propose to first uplift the 2D semantics into layered 3D semantics with
a structure similar to multi-plane images (MPI) for RGB images~\cite{szeliski1998stereo,zhou2018stereo}.
We call this structure \checkthis{\emph{lifted semantics}}.
Unlike MPIs, to relax the memory requirements and for translation efficiency, our \checkthis{lifted semantics} use a hybrid representation with a small set of semantic layers and a larger set of transparency layers.
We convert the \checkthis{lifted semantics} to layered features, which we refer to as \emph{layered appearance}, and combine them with the transparency layers.
Finally, we project the resulting appearance features onto the target views and convert them to RGB images with a small network.
The late fusion of the \checkthis{lifted semantics} is key to the quality of our results.

We perform extensive experimental analysis on three different multi-view datasets: CARLA~\cite{dosovitskiy2017carla}, Cityscapes~\cite{cordts2016cityscapes}, and Virtual-KITTI-2~\cite{cabon2020virtual}.
We show both qualitatively and quantitatively that our approach, which compares favorably with strong baseline techniques, produces novel-view images that are geometrically and semantically consistent.
In addition, we also demonstrate that we can estimate high-quality depth information from single-view semantics.

\section{Related Works}
\vspace{-2mm}

Novel view synthesis (NVS) has a rich history that predates the deep learning era~\cite{chen1993view,zitnick2004high}.
Traditional methods tackle the challenge of generating pixels for unseen viewpoints with proxy geometry~\cite{buehler2001unstructured,debevec1996modeling}, or with a significant number of input images~\cite{chaurasia2011silhouette,CDSD13}.
Thanks to learned priors, impressive results with as little as two input images and no additional information are also possible~\cite{choi2019extreme,zhou2018stereo,srinivasan2019pushing,flynn:2019:deepview}.
A key regularization technique at the core of many of these methods is a scene representation consisting of a set of fronto-parallel layers that can be merged down into the target view, after appropriate warping.
This approach, which relates to representations proposed by earlier methods~\cite{szeliski1998stereo}, is dubbed Multi-Plane Images (MPI)~\cite{zhou2018stereo} and is also central to the success of our method, albeit with significant modifications.
The most recent NVS works go even further and show single-image NVS by learning to predict single-image depth~\cite{wiles2019synsin,contvs}, or by extending MPI imaging to single input image~\cite{single_view_mpi}.

While powerful, all of these NVS methods require an input RGB image.
To better leverage their creative agency, users can also edit a simplified representation of a scene or object, which can then be converted to an RGB image.
For instance, sketches can be turned into photorealistic pictures~\cite{chen2009sketch2photo,pix2pix}.
A scene-level representation that is particularly flexible is afforded by semantic maps.
Indeed, a number of works using both traditional tools~\cite{johnson2006semantic} and deep learning~\cite{chen2017photographic,pix2pixhd,pix2pix} produce impressive semantic-to-RGB results.
An example of the engagement this type of technology can enable is offered by the method by Park et al.~\cite{spade}, which takes a hand-drawn semantic map and produces an RGB image. Within months of publication, more than 500,000 images were created by web users~\cite{gauganBlog}.

Image-to-image translation offers more control over the content generated, but unlike NVS methods, it does not allow to modify the viewpoint.
We propose to combine the advantages of the two lines of work.
Our method takes a semantic image as input and produces photorealistic images from novel viewpoints. 
However, compared with the sequential combination of existing methods, our strategy better leverages the information the semantic maps offer: a robust representation of the scene structure.
We discuss this in more detail in the following section.

\section{Approach}\label{sec:approach}
\vspace{-2mm}

We present \textit{Generative View Synthesis} (GVS), a method that takes a single 2D semantic map and generates photorealistic images from novel viewpoints.
One way to tackle this problem is by a straightforward combination of existing techniques, that is, 2D semantics to 2D image conversion in the reference view~\cite{spade}, followed by monocular novel view synthesis~\cite{contvs, appflow, wiles2019synsin} to generate the target views.
However, this na\"ive approach fails at preserving some of the structures observed in the semantic input, as can be observed in Figure~\ref{fig:teaser}.
This is because the semantic map's strong cues about the layout of the scene are lost in the early conversion to RGB.
In contrast, GVS carries the semantic information forward, and only converts it to RGB after its projection onto the target viewpoint.
This results in photorealistic target views that are geometrically consistent and that better preserve the structures in the input semantics. 

\noindent \textbf{Approach Overview.}
Formally, GVS takes a 2D semantic map \s$^r \in \mathbb{R}^{n \times l}$ in the reference view $r$, where $n$ is the number of pixels and $l$ the number of labels. It also takes the relative camera pose transformation from source to target view $\theta_{r \to t} \in SE(3)$. 
The output is a image $I^t \in \mathbb{R}^{n \times 3}$ in the target view $t$.
Additionally, we use an image $Q \in \mathbb{R}^{n \times 3}$ to control the style of the generated images.

We train a network to convert the 2D input semantics \s$^r$ to \mps$^r$, a layered semantics representation that we call \checkthis{\emph{MPI semantics}}.
\checkthis{MPI semantics} are inspired by multi-plane images (MPIs)~\cite{szeliski1998stereo,zhou2018stereo}, which represent a 3D scene with a stack of 2D layers positioned at $m$ depth levels $(d_1, d_2, ..., d_m)$. 
We represent the \checkthis{MPI semantics} with $m$ layers: \mps$^r \in \mathbb{R}^{n \times l \times m}$.
Each layer contains semantic labels at each pixel, at that layer's depth, and the pixel transparency $\alpha \in \mathbb{R}^{n \times m}$. MPI representations~\cite{zhou2018stereo,szeliski1998stereo} are a widely used
representation for novel view synthesis because their 2D nature allows for well-studied and powerful processing techniques such as convolutional neural networks (CNN).

One could project the MPI semantics onto the target views and independently translate each of the target-view semantics into RGB images. 
This would enforce geometric consistency, but it would not guarantee the appearance of the output images to be consistent.
Therefore, we propose a translation scheme that converts the \checkthis{MPI semantics} to layered 3D appearance features first.
We then project the layered 3D appearance features onto the target views and convert them to the target RGB images.
We demonstrate that this approach results in multi-view-consistent images and preserves the input semantic structures.
Figure~\ref{fig:pipeline} (top) shows an overview of our approach:
we uplift the 2D semantics using a \textit{semantics uplifting network} (SUN) and translate the lifted semantics to layered appearance features using a \textit{layered translation network} (LTN).
We then project the layered appearance features to novel views and convert projected features using an \textit{appearance decoder network} (ADN) to generate target images. 

We find that the use of MPI semantics, however, is intractable because of the large memory footprint of the layered translation network (LTN).
A faithful approximation of any content in 3D space, in fact, requires $m >$ 32  MPI layers, making the semantics-to-appearance MPI translation infeasible---recall that we have $l$ labels for each layer, and each pixel.
Therefore, we propose a hybrid layered representation.
Specifically, we learn only $k<m$ semantic layers and the full set of $m$ transparency layers, which we then combine by learning an association function.
We call our overall network for generative view synthesis \emph{GVSNet}.

\subsection{Semantics Uplifting Network}
\label{sec:uplifting_net}
\vspace{-2mm}

We first uplift the input 2D semantics to layered 3D semantics using a 2D CNN, which we refer to as \emph{Semantics Uplifting Network}~(SUN). 
As outlined above, instead of converting the semantics to MPI-semantics, we propose to use a hybrid representation.

\noindent \textbf{Hybrid Layered 3D Semantics Representation.}
Representing semantic information at each of the $m$ depth layers 
of the MPI has a memory footprint in the order of $O(m \times n \times l)$.
In practice, however, the 3D scene is mostly empty, causing each of the layers in the MPI to be sparse.
Therefore, we propose to represent the layered 3D semantics with fewer layers, $k<m$, which we call \emph{lifted semantics}, \ls$^r \in \mathbb{R}^{n \times l \times k}$. 
For this representation, we use the input 2D semantics as
the first layer and predict the remaining layers using the SUN network.
In practice, we observe that $k=$3 layers (including the input semantics) suffice. 
On the other hand, we do not compress the transparency, $\alpha$, to fewer layers, as it serves as a proxy to the scene geometry and it only requires a scalar value for each pixel in each layer. 
That is, we represent transparency $\alpha$ with the original number of MPI layers $m$.
The transparency layers are also shared with the layered 3D appearance features. 

\begin{figure}[t!]
\centering
\includegraphics[width=0.95\textwidth]{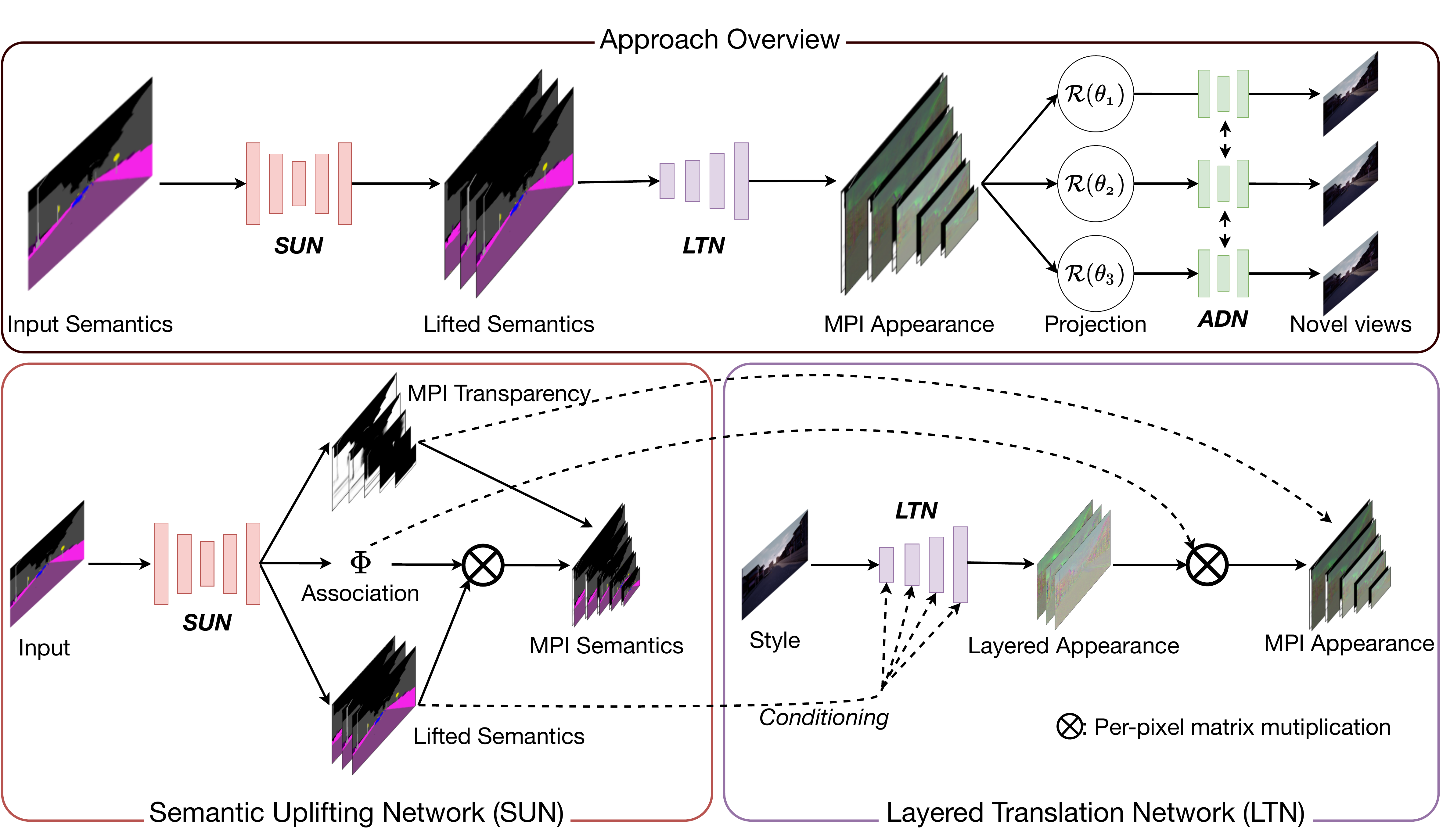}
\caption{\textbf{GVSNet Overview.} [Top] Approach overview illustrating the lifting of semantics to layered 3D representation using Semantic Uplifting Network (SUN) and then translating to MPI appearance with Layered Translation Network (LTN). We then project the MPI appearance onto target views and use Appearance Decoder Network (ADN) to generate images. [Bottom] Illustration of different representations and operations in SUN and LTN.}
\label{fig:pipeline}
\end{figure}

\begin{figure}[h!]
    \centering
    \input{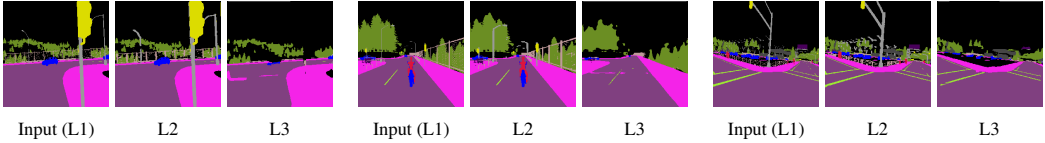}
  \caption{\textbf{Lifted Semantics.} L1, L2 and L3 correspond to the first, second and third lifted semantic layers respectively. We observe that the lifted semantic layers loosely correspond to occlusion layers, where the later layers capture the content occluded by the earlier layers, see L3 images. Another interesting observation is that, at mid levels~(L2), the network \textit{dilates} thin structures (see the poles).}
  \label{fig:layered_sem_rendering}
\end{figure}

Because of the mismatch in the number of layers, we need to estimate an association map $\Phi \in \mathbb{R}^{n \times k \times m}$ to convert the lifted semantics \ls$^r$ to MPI semantics \mps$^r$.
At each pixel $p$, we can convert the lifted semantics \ls$^r_p \in \mathbb{R}^{l \times k}$ into MPI semantics \mps$^r_p \in \mathbb{R}^{l \times m}$ representation with the column-normalized association matrix $\Phi^*_p \in \mathbb{R}^{k \times m}$: \mps$^r_p = \tilde{\mathcal{S}}^{r}_{p} \, \Phi^*_p$.
Figure~\ref{fig:pipeline} (bottom-left) illustrates the SUN network that takes 2D semantics in the reference view as input and predicts the lifted semantics \ls$^r$, the MPI transparency $\alpha$, and the association map $\Phi$. 
Figure~\ref{fig:layered_sem_rendering} shows sample lifted semantics layers. We visually observe that the lifted semantics layers roughly correspond to occlusion layers, where the farther layers capture the content occluded by the closer layers.

\subsection{Layered Translation Network}
\label{sec:layered_translation}
\vspace{-2mm}

With the the MPI semantics \mps$^r$, we can render the semantics into the target views $\mathcal{S}^t$, and use any of the recent conditional image generation networks~\cite{spade} to generate target-view images. 
We empirically observe that independent translations of semantics to RGB images can result in inconsistent results---even when the structure is consistent, the corresponding texture may vary across views.
To remedy this, we estimate layered 3D appearance features and directly translate the lifted semantic layers to appearance feature layers.
Differently put, we carry forward for as long as possible a view-independent representation of the scene.

State-of-the-art image-to-image translation networks use large generative adversarial networks (GAN), making it infeasible to translate MPI semantics to MPI appearance features. 
This is the very reason why we use SUN to predict lifted semantics \ls$^r$ instead of MPI semantics \mps$^r$.
We use a \emph{Layered Translation Network} (LTN) to convert $k$ lifted semantics layers \ls$^r$ to $k$-layered appearance features \la$^r \in \mathbb{R}^{n \times f \times k}$, where $f$ denotes the appearance feature dimensionality.
LTN is a modified version of SPADE~\cite{spade}.
Like the original SPADE, LTN can also take a style image as input to enable easy manipulation of the appearance of the generated images.
However, SPADE takes 2D semantics as input and generates 2D image, while LTN takes multi-layer inputs and produces multi-layer outputs.
We then use the same MPI transparency layers $\alpha$ and association map $\Phi$, estimated with SUN, to convert the $k$-layered appearance features to MPI appearance features with $m$ layers.
That is, at each pixel $p$: \mpa$^r_p =$\la$^r_p\Phi^*_p$.
Figure~\ref{fig:pipeline} (bottom-right) illustrates the LTN network.

\subsection{Appearance Decoder Network}
\label{sec:view_projection}
\vspace{-2mm}
We render the MPI appearance features \mpa$^r$ into target-view appearance $A^t \in \mathbb{R}^{n \times f}$ for a given target-view $t$. 
We then train a small CNN called \textit{Appearance Decoder Network} (ADN) that converts target-view appearance to the final target image $I^t$. 
As an alternative, one could directly estimate color MPI using LTN and then just render the color MPI onto target-view to obtain a target image. 
We empirically observe that projecting high-dimensional ($f$-dimensional) MPI features can result in better target views in comparison to projecting color MPI. 
This is because high-dimensional, per-pixel features help mitigate some of the artifacts that arise from the discrete nature of the MPI planes. The high-dimensional features capture neighborhood pixel information and can provide more contextual information to ADN to deal with possible MPI rendering artifacts.

\noindent \textbf{Loss Functions.}
The overall GVSNet, illustrated in Figure~\ref{fig:pipeline}, has three main sub-networks: Semantics uplifting network (SUN), Layered translation network (LTN) and Appearance decoder network (ADN).
To train GVSNet, we use a weighted sum of target segmentation loss $\mathcal{L}_{sem}$, depth loss $\mathcal{L}_{dep}$, target color loss $\mathcal{L}_{col}$ and GAN loss $\mathcal{L}_{gan}$:
\begin{equation}
    \mathcal{L} = \lambda_{0}\,\mathcal{L}_{sem} + \lambda_{1}\,\mathcal{L}_{dep} + \lambda_{2}\,\mathcal{L}_{col} + \lambda_{3}\,\mathcal{L}_{gan}.
    \label{eqn:loss}
\end{equation}

$\mathcal{L}_{sem}$ denotes the negative log-likelihood loss for the predicted semantics in the target view. We project the MPI semantics to target-view semantics using standard MPI rendering~\cite{zhou2018stereo} that involves homography transformation and alpha composition.
Refer to~\cite{zhou2018stereo} for more details on MPI rendering. 
We then use negative log-likelihood loss $\mathcal{L}_{sem}$ between ground-truth (GT) semantics and predicted semantics in the target views.
For the depth loss, we first compute per-pixel inverse depth $D^r$ in the reference view from the predicted MPI transparency $\alpha$ by performing back-to-front alpha composition on the inverse-depth values of each plane at the given pixel. 
Refer to the supplementary for more details on the inverse depth computation from transparency. 
Then the depth loss $\mathcal{L}_{dep}$ is the L1 distance between the GT inverse depth and the computed inverse depth. 
The target color loss and GAN lossses, include the same losses as in SPADE~\cite{spade}, that is, perceptual loss on generated color image, discriminator feature reconstruction loss and GAN losses. The main difference to SPADE~\cite{spade} is that we use these loss functions on generated \emph{target} image with respect to the GT target image.

We train GVSNet in two stages. In the first stage, we pre-train SUN with the target segmentation and depth losses. 
In the second stage, we train LTN and ADN with the target color loss, while keeping the SUN fixed.
We use this two-stage training because the entire network training does not fit on NVIDIA GTX-2080-Ti GPUs, which is what we use for training. 
However, we could conceptually train GVSNet end-to-end, because all the components are differentiable. 
The SUN network is a composed of two parts: backbone and three prediction heads~(for $\alpha$, $\tilde{\mathcal{S}}^{r}$ and $\Phi$). The backbone is a UNet~\cite{ronneberger2015u} style encoder-decoder network with $7$ encoding and decoding stages. The prediction heads are convolutional blocks that share their first 3 layers. ADN is a light CNN with 5 encoder-decoder layers. The LTN network is a SPADE~\cite{spade} network with 7 SPADE-Residual blocks and UpSampling layers. For our experiments, we used $k=3$ lifted semantics layers, $m=32$ MPI planes and $f=20$ appearance features per pixel. We implemented our model in PyTorch~\cite{paszke2019pytorch} and use the Adam~\cite{kingma2014adam} optimizer for training. More details about the training and network architectures are given in the supplementary material. 

\section{Experiments}
\vspace{-2mm}

\begin{figure}[t!]
    \centering
    \input{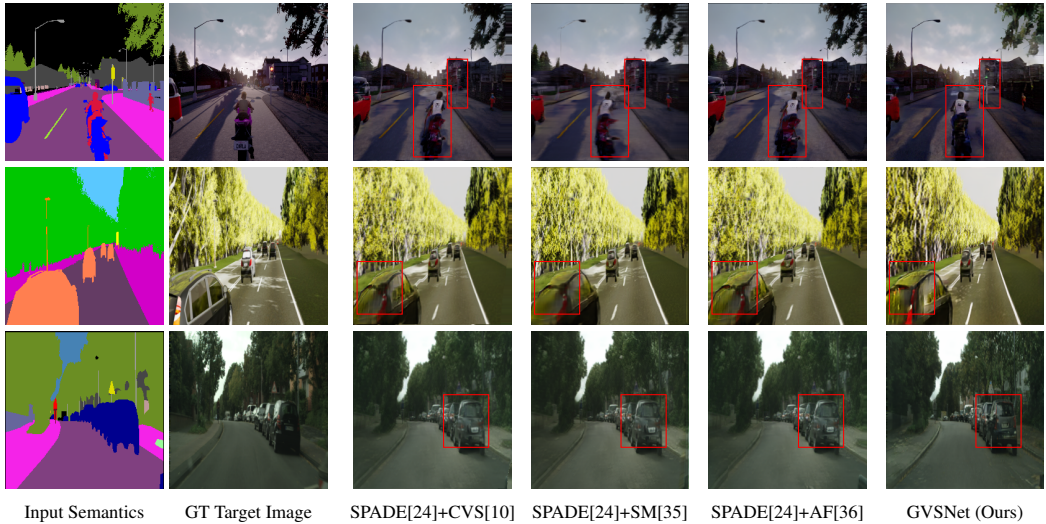}
    \caption{\textbf{Sample Visual Results} showing generated novel-view images on CARLA (top), Virtual-KITTI-2 (middle), and Cityscapes (bottom) images.}
    \label{fig:results}
\end{figure}

\noindent \textbf{Datasets.} GVSNet is fully supervised and thus requires datasets providing two or more views for each scene. 
It also needs semantic segmentation annotations for at least one view. 
Given these constraints, we perform experiments on three different datasets: CARLA~\cite{dosovitskiy2017carla}, Virtual-KITTI-2~\cite{cabon2020virtual} and Cityscapes~\cite{cordts2016cityscapes}.
We use a pair of cameras to train GVSNet, treating one of the images as the input view and another one as the target view. In CARLA, we sample pairs from a set of cameras arranged along $x$-(left-right) and $z$-(forward) axes. For Virtual-KITTI-2 and Cityscapes, we use stereo pairs with a horizontal baseline.
For CARLA and Virtual-KITTI-2, following the practice in~\cite{spade}, we use the color image from the input camera as a style guidance when generating the target view. In order to make our results comparable to SPADE~\cite{spade}, we do not use style input images in Cityscapes experiments. For the Cityscapes dataset, the ground truth (GT) semantic segmentation is only available for the left camera. We used a pre-trained semantic segmentation network~\cite{semantic_cvpr19} to generate semantics for the right camera images. As there is no ground truth depth in Cityscapes, we computed depth maps by training the DPSNet~\cite{im2019dpsnet} in a self-supervised manner. Moreover, as SPADE~\cite{spade} is trained with both semantic and instance segmentation masks on Cityscapes, we also use instance masks by concatenating the input view instance mask with lifted semantics. Because instance masks are available only for the input view, we create target view instance mask by warping the input view masks to the target view, using depth estimated by DPSNet~\cite{im2019dpsnet}. In all of our experiments we use images at a resolution of $256\times 256$ pixels.

\noindent \textbf{Evaluation Metrics.} 
The results of a GVS system should have two properties: 
1. \emph{Semantic Preservation}: The generated image should retain the semantic structures of the input semantics; and
2. \emph{Photorealism}: The generated target images should be photo-realistic.
To measure semantic preservation, we apply a semantic segmentation network (DeeplabV3+~\cite{deeplabv3p}) on the synthesized images and compare its output with the GT semantics in the target view.
Specifically, we report mean class accuracy and mean Intersection over Union (IoU) as segmentation metrics.
To measure photo-realism, we report the Fréchet Inception Distance (FID) score~\cite{heusel2017gans} and Perceptual Distance (PD)~\cite{zhang2018perceptual} metric, which measure the distance between generated target-view images and GT images in VGG~\cite{vgg} feature space.

\begin{table}[h]
\centering
\resizebox{\textwidth}{!}{%
\small
\begin{tabular}{lcccccccccc}
\toprule
                   & \multicolumn{4}{c}{CARLA~\cite{dosovitskiy2017carla}} & \multicolumn{4}{c}{Virtual-KITTI-2~\cite{cabon2020virtual}} & \multicolumn{2}{c}{Cityscapes~\cite{cordts2016cityscapes}} \\
\cmidrule(lr){2-5}
\cmidrule(lr){6-9}
\cmidrule(lr){10-11}

Method             & Cls. Acc.~$\uparrow$        & IoU~$\uparrow$       & PD~$\downarrow$ & FID~$\downarrow$   &  Cls. Acc.~$\uparrow$        & IoU~$\uparrow$       & PD~$\downarrow$ & FID~$\downarrow$      & PD~$\downarrow$ & FID~$\downarrow$  \\ \midrule
GVSNet (Ours)   & \textbf{74.34} & \textbf{66.43} & \textbf{1.74}  & \textbf{62.06} & \textbf{77.13} & \textbf{69.62} & \textbf{2.08}                & \textbf{36.21} & \textbf{2.76} & \textbf{48.72} \\
\midrule
SPADE~\cite{spade} + SM~\cite{zhou2018stereo}         & 69.93  & 60.82  & 1.95   & 75.81  & 74.84 & 64.71 & 2.19             & 41.61  & 2.82 & 60.71  \\
SPADE~\cite{spade} + CVS~\cite{contvs}     & 66.84  & 57.29   & 1.88  & 69.24  & 76.23 & 67.73 & 2.12                 & 37.79  & 2.80 & 57.46 \\
SPADE~\cite{spade} + AF~\cite{appflow}     & 66.15  & 56.45  & 1.92 & 76.89 & 76.81 & 68.66 & 2.15 & 40.95  & 2.83&57.15 \\ 
\midrule
Target GT Images             & 77.47  & 69.67   & -   & - & 83.58 & 75.39 & - & - & - & -     \\ \bottomrule    
\end{tabular}%
}
\caption{\textbf{Comparisons to Baselines.}
Semantic segmentation (Class Accuracy and IoU), FID~\cite{heusel2017gans} and Perceptual Distance (PD) evaluations on different datasets for GVSNet (ours) along with SPADE+X baseline techniques, that first perform semantics-to-RGB conversion followed by monocular NVS.}

\label{table:carla}

\end{table}

\begin{figure}[h!]
    \centering
    \includegraphics[width=\textwidth]{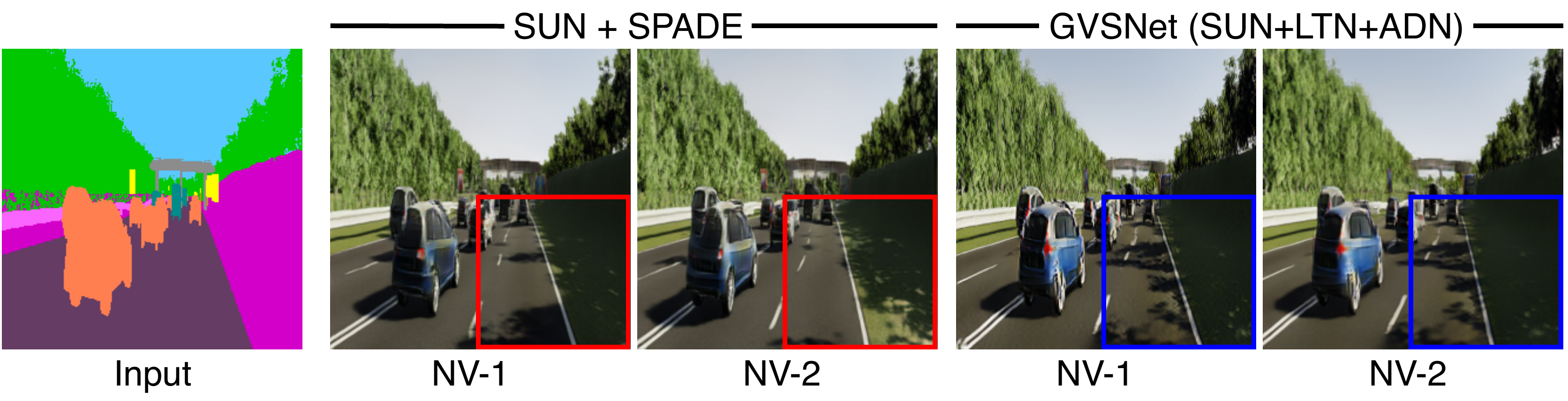}
    \caption{\textbf{Consistency across views.}
    We show sample generated images in two novel views (NV-1 and NV-2) using our full GVSNet model along with those obtained with SUN+SPADE. SUN+SPADE model first estimates novel-view semantics followed by semantics-to-RGB translation with SPADE~\cite{spade}. Results show that view-independent translation results in inconsistency across the views and the layered translation (LTN) is important to generate view-consistent images. As can be observed in the red boxes, SUN+SPADE changes texture of the same physical area between novel views, while GVSNet gives consistent renderings~(see blue boxes).}
    \label{fig:inconsistency}
\end{figure}

\noindent \textbf{Comparisons to Baselines.}
As this is the first work to tackle GVS, there is no existing baseline technique against which we can directly compare. 
To properly evaluate our approach, then, we propose sensible baseline methods based on the adaptation and combination of state-of-the-art methods.
Specifically, we use a pipeline that converts 2D semantics to an RGB image with SPADE~\cite{spade} and applies monocular NVS techniques to render novel views.
For the latter task, we pick Stereo-magnification (SM)~\cite{zhou2018stereo} adapted to single-view NVS, Continuous View Synthesis (CVS)~\cite{contvs}, and Appearance Flow (AF)~\cite{appflow}.
We refer to these baselines as `SPADE+X' where `X' could be `SM', `CVS' or `AF'. 
Table~\ref{table:carla} shows the quantitative results.
Results show that GVSNet is consistently superior to the SPADE+X techniques across different metrics and datasets.
This demonstrates that GVSNet can better preserve the semantic structures of the input while generating more realistic images. 
Figure~\ref{fig:results} shows sample visual results.
The visual results in Figure~\ref{fig:teaser} (please, play the animation), Figure~\ref{fig:results}, and Figure~\ref{fig:inconsistency} further validate that GVSNet better preserves semantic structures and geometric consistency.
We show more visual results in the supplementary material.

\noindent \textbf{Ablation Study.}
GVSNet comprises several computational blocks, each critical to its success.
To confirm this, we perform an ablation study and evaluate the impact of swapping out parts of it.
One basic test is to only use the semantic uplifting network (SUN) to synthesize 2D semantics in the target view, and use SPADE~\cite{spade} to convert  it to RGB image.
We refer to this model as `SUN+SPADE'.
To evaluate the importance of $f$-dimensional layered appearance features ($f$=20) as opposed to just estimating 3-dimensional color, we experiment with a model that translates layered semantics to layered color images. We refer to this variant as `SUN+LTN', as it does not use ADN.

\begin{table}[h]
\centering
\scriptsize
\begin{tabular}{lcccc}
\toprule
Method             & Class Acc.~$\uparrow$        & IoU~$\uparrow$       & PD~$\downarrow$ & FID~$\downarrow$   \\ \midrule
\multicolumn{2}{l}{\textit{GVSNet Variations}} & & & \\
SUN + SPADE~\cite{spade}  & 72.92         & 65.52          & 1.75     & 68.96 \\ 
SUN + LTN           & 71.90  & 63.12 & 1.83 & 69.46\\
SUN + LTN + ADN (Full model)   & \textbf{74.34} & \textbf{66.43} & \textbf{1.74}  & \textbf{62.06} \\
\bottomrule
\end{tabular}%
\caption{\textbf{Ablative Studies on GVSNet.} 
Semantic segmentation (Class Accuracy and IoU), FID~\cite{heusel2017gans} and Perceptual Distance (PD) evaluations on CARLA~\cite{dosovitskiy2017carla} with different variations of our GVSNet. Results show that all the component networks of SUN, LTN and ADN are important.}
\label{table:ablation}
\end{table}

Ablation results shown in Table~\ref{table:carla} indicate that all the three component networks (SUN, LTN and ADN) are important.
Performance drops considerably if we use LTN to directly translate to color MPI (SUN+LTN vs. SUN+LTN+ADN). 

We observe that SUN+SPADE, that applies SPADE~\cite{spade} on projected 2D semantics in novel views, can generate photo-realistic novel-view images.
However, we also notice that SUN+SPADE produces inconsistent images across multiple target views as image translation is applied independently on each target view semantics. Figure~\ref{fig:inconsistency} shows a sample visual result indicating much better view-consistency with GVSNet when compared to SUN+SPADE model. We argue that, in the context of content generation, geometric or appearance inconsistency among different views is more detrimental than a slight perturbation of the semantic information, when this is common to all the synthesized views. We present more results in the supplementary material.

\begin{figure}[htb!]
  \input{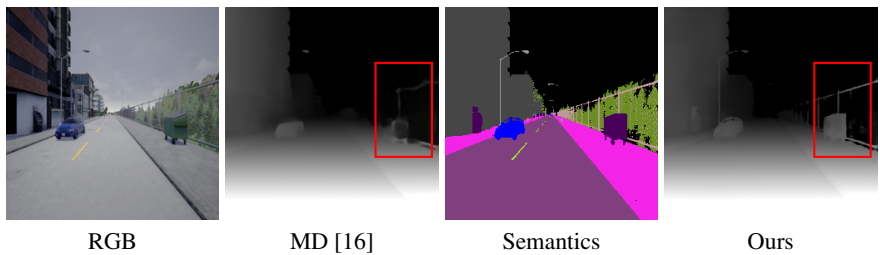}
    \caption{\textbf{Depth Estimation} We can obtain relative depth using the estimated MPI transparency from a single input semantics. Also shown is the result obtained with MonoDepth~\cite{monodepth} (MD) that uses RGB image as input. Results show that our model can estimate better geometric structures compared to MD.}
  \label{fig:depth}
\end{figure}

\noindent \textbf{Depth Estimation.}
Our proposed method infers the 3D scene structure from the input semantics. Since we estimate transparency $\alpha$ at each MPI plane, we can convert the MPI transparency into depth by performing alpha composition on the depth values of the MPI planes (see supplementary for details). The ability to generate realistic depth maps from semantic images can also simplify content creation tasks. In this section, we compare the quality of depth maps learned with our method against established depth estimation approaches~\cite{monodepth}. Figure~\ref{fig:depth} shows a sample depth estimation with our GVSNet that \textit{only} takes 2D semantics as input. For comparison, we also show depth estimation from the RGB image using MonoDepth~\cite{monodepth} (MD), which is trained in a fully-supervised manner using GT depths.  Results indicate that our depth estimates are more accurate at thin structures. This is further confirmation of the strong structural cues offered by the semantics. We present more qualitative and quantitative results in the supplementary material. 

\subsection{Applications}\label{sec:applications}
Generative view synthesis lends itself to a number of interesting applications. Here we focus on one such example: semantic editing.

\begin{figure}[htb!]
    \centering
    \input{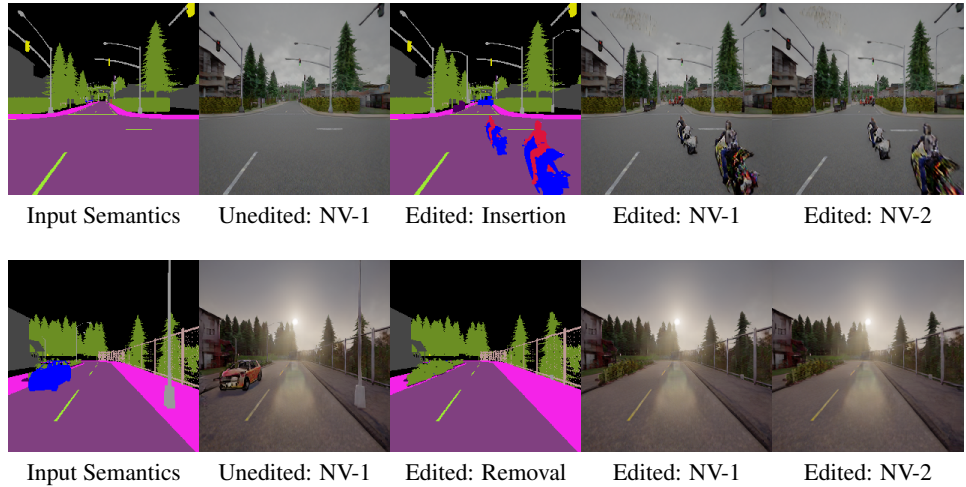}
    \caption{\textbf{Semantic Editing.} The two rows show different scenes rendered in color before and after and manipulation. Unedited: NV-1 is the rendering of the input semantics at the input camera pose. Edited: NV-1 and NV-2 are novel views of the edited scene at the input camera and a novel camera pose, respectively. Unlike semantics manipulation with image-to-image translation methods, GVSNet allows to manipulate the input semantics and generate consistent novel-views of the edited scene. Results with both object insertion (top) and object removal (bottom) show that GVSNet can generate realistic novel-views also on manually edited semantics.}
    \label{fig:scene editing}
\end{figure}

\noindent \textbf{Semantic Editing.}
One of the main advantages of GVS is to simplify multi-view content generation. GVS allows the effect of manipulating the input 2D semantic map to be seamlessly propagated to all the novel views. Editing a semantic map by simply pasting an object from another, for instance, is arguably easier than directly editing RGB pixels, which requires accounting for lighting conditions explicitly, adding realistic textures, etc. Similarly, despite the success of recent inpainting methods~\cite{liu2018image}, removing objects from RGB images while accounting for the scene context requires advanced skills. Semantic manipulation with image-to-image translation methods such as SPADE~\cite{spade}, is limited to static 2D images. Thanks to our method, simple object addition to/removal from the 2D semantic maps seamlessly translates to photorealistic images of novel views, as shown in Figure~\ref{fig:scene editing}.  
\vspace{-2mm}
\section{Conclusion}
\vspace{-2mm}

In this work, we propose Generative View Synthesis Network (GVSNet), which produces photo-realistic novel-view images from only a single 2D semantic map.
We demonstrate that the simple application of existing techniques for this problem yields inadequate results. 
Our key insight is to leverage the structural information in the input semantics by uplifting the 2D semantics to layered 3D semantics.
Further, we carry the structural 3D semantic information forward with a layered semantics-to-RGB translation network.
Comprehensive experimental analysis on three different datasets demonstrate the potential of GVSNet in generating geometrically consistent novel-view images, while preserving the structures in the input semantics.

\section*{Acknowledgement}
\vspace{-2mm}
This project was partially funded by the BMBF project VIDETE(01IW1800). We thank the SDS department at DFKI Kaiserslautern, for their support with GPU infrastructure.

\small

\bibliographystyle{plain}
\bibliography{ms}

\clearpage
\begin{center}
\part{\textbf{\large Supplementary Material}}
\end{center}
\section{Additional Experimental Results}
In this supplementary, we provide additional details on depth estimation and evaluation(in section~\ref{sec:depth}), evaluation with different style images(in section~\ref{sec:style}), network architecture(~\ref{sec:architectures}) and additional training details(in section~\ref{sec:training}).

\subsection{Visual Results}
\begin{figure}
  \begin{center}
    \includegraphics[width=\textwidth]{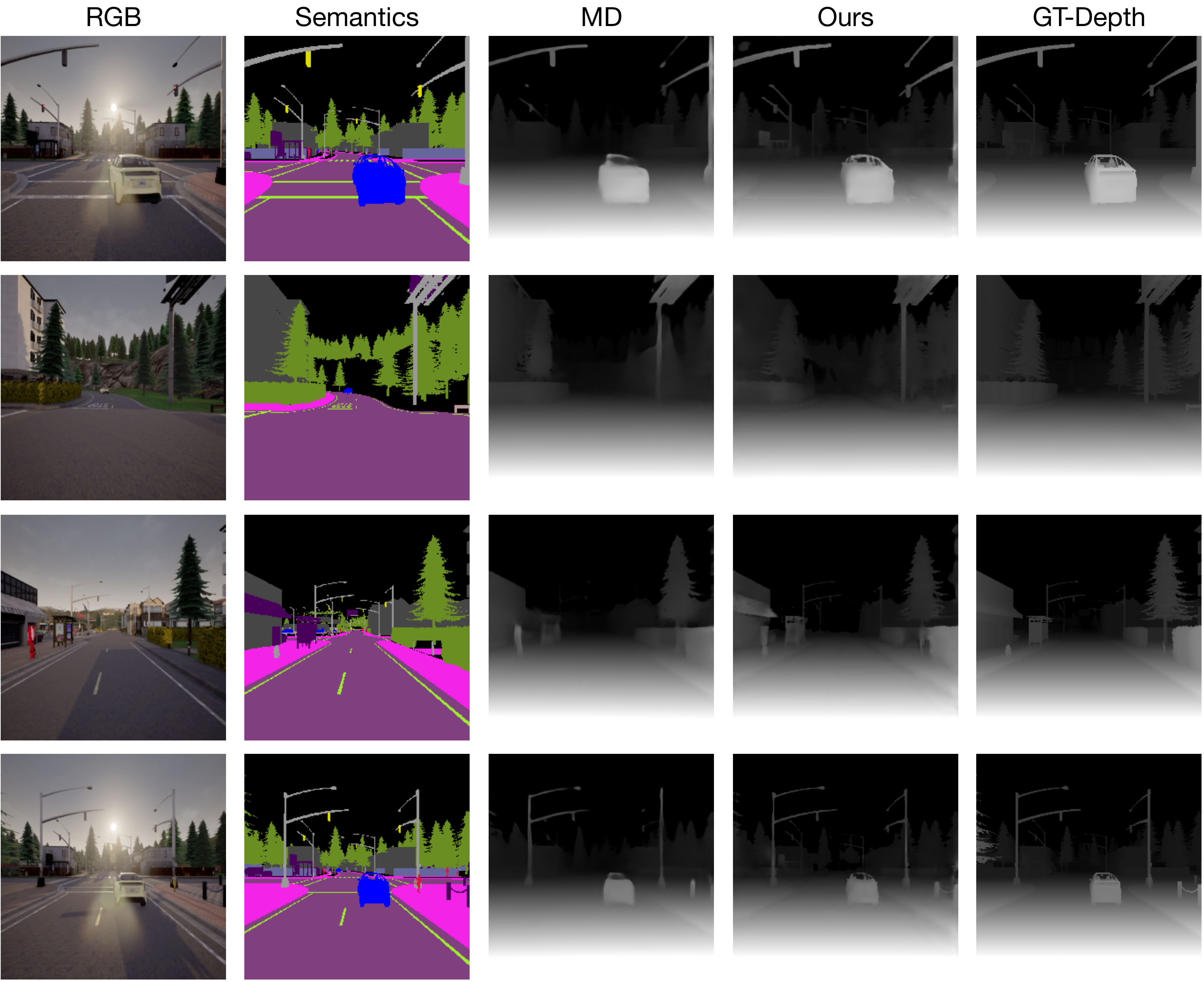}
  \end{center}
  \caption{\textbf{Depth Estimation Results.} Here we present more visual depth estimation results from our SUN model and MonoDepth~(MD) model trained with ground-truth depths.}
  \label{fig:depth_sup}
\end{figure}

More visual results on depth estimation are included in Figure~\ref{fig:depth_sup}.

\subsection{Depth Estimation and Evaluation}\label{sec:depth}
\noindent \textbf{Extracting Depth Maps from Multi-Plane Transparencies.}
Our Semantic Uplifting Network~(SUN) estimates multi-plane transparencies $\alpha$, in the reference camera. Here we show a simple way to convert~$\alpha \in R^{n\times m}$ in to a depth map $\hat{Z}\in R^{n}$, where $n$ is the number of pixels and $m$ is the number of planes in the MPI representation. Suppose the MPI planes are located at distances $\{d_{1}, d_{2}, \dots, d_{m}\}$ and are fronto-parallel to the reference camera. The depth value at a pixel $p$, $\hat{Z}(p)$, is computed by alpha compositing the depth of the MPI planes with the alpha values at pixel $p$, as follows. 

\begin{equation}
\hat{Z}(p) = \displaystyle\sum_{i=1}^{m}{ \begin{bmatrix} d_{i} \,  \alpha(p,i) \, \begin{bmatrix} \displaystyle\prod_{j=1}^{i-1}{(1-\alpha(p,j))} \end{bmatrix} \end{bmatrix}} 
\label{eqn:pseudo_depth}
\end{equation}

\noindent \textbf{Depth Evaluations.}
In Table~\ref{table:depth}, we show accuracy of the depths estimated by our SUN method and compare it against a baseline network trained with ground truth depths. The baseline network uses an RGB input while our method uses semantics as input. The baseline network is taken from the MonoDepth~\cite{monodepth}; it is an encoder decoder network with ResNet-18 backbone. The network can be trained in a self-supervised manner, however for a fairer comparison, we train the network using ground-truth depth.

We evaluated these networks following traditional depth accuracy metrics~\cite{ummenhofer2017demon}: \textit{scale invariant depth} (SC\_Inv), \textit{relative depth} (L1\_Rel) and \textit{inverse depth} (L1\_Inv) metrics. Using only semantics as input, which is devoid of photo-metric details, our network produces favourable depth estimations when compared to the baseline. Our method outperforms the MonoDepth network on SC\_Inv and L1\_Inv metrics. However, on L1\_Rel, our method under-performs compared to MonoDepth. We believe this is due to the fact that MPI planes are distributed by sampling the inverse depth linearly. This results in having few planes far from the camera. Nonetheless, for the view synthesis tasks it is desirable to have higher accuracy at closer ranges.

\subsection{Evaluations with Different Style Images.}\label{sec:style}
\vspace{-2mm}
As can be seen in the Figure-1 of the main paper and in the supplementary video, our method can produce novel-views with different styles, as dictated by the given style images. In image-to-image literature~\cite{spade}, it is customary to use the target color image as a style during evaluation. Following this custom, we use the color image from the source camera as a style guidance for the novel-view image generation. In order to verify the efficacy of method under arbitrary styles, we perform an additional test where our method is not fed the input image as style, rather a random frame the same sequence is used as style input. In Table~\ref{table:carla_style}, we show semantic evaluation results using this procedure. The results show that our method still outperforms the baselines under this setting of using different style image. 

\begin{table}[h!]
\centering
\scriptsize
\begin{tabular}{@{}lllllll@{}}
\toprule
\multirow{2}{*}{Method} &
  \multirow{2}{*}{Input} &
  \multirow{2}{*}{Supervision} &
  \multirow{2}{*}{SC\_Inv} &
  \multirow{2}{*}{L1\_Rel} &
  \multirow{2}{*}{L1\_Inv} &
  Depth range \\
           &     &              &       &       &       & (in meters) \\ \midrule
Ours       & Sem & Sem + Depth  & \textbf{0.200} & 0.129 & \textbf{0.006} & 1 - 100     \\
Mono-Depth & RGB & Depth        & 0.225 & \textbf{0.125} & 0.008 & 1 - 100     \\ \midrule
Ours       & Sem & Sem +  Depth &\textbf{ 0.252} & 0.184 & \textbf{0.006} & 1 - 200     \\
Mono-Depth & RGB & Depth        & 0.269 &\textbf{ 0.155} & 0.008 & 1 - 200     \\ \midrule
Ours       & Sem & Sem + Depth  & 0.\textbf{331} & 0.395 &\textbf{ 0.006} & 1 - 1000    \\
Mono-Depth & RGB & Depth        & 0.358 & \textbf{0.268} & 0.007 & 1 - 1000    \\ \bottomrule
\end{tabular}%
\caption{\textbf{Depth accuracy evaluation on CARLA~\cite{dosovitskiy2017carla} dataset.} Depth maps generated by our SUN model and monocular depth estimation network from~\cite{monodepth}. SC\_inv, L1\_Rel and L1\_Inv stand for scale invariant, L1 relative and L1 inverse depth metrics, respectively.}
\label{table:depth}
\end{table}

\begin{table}[h!]
\centering
\scriptsize
\begin{tabular}{lcc}
\toprule
Method             & Cls. Acc. $\uparrow$     & Mean IoU $\uparrow$ \\ \midrule
\textit{GVSNet Variations} & & \\
GVSNet (Full Model)     & \textbf{73.16} & \textbf{65.40}   \\
SUN+SPADE~\cite{spade} & 70.95  & 63.73   \\ \hline
SPADE~\cite{spade} + SM~\cite{zhou2018stereo}  & 67.26 & 58.71   \\
SPADE~\cite{spade} + CVS~\cite{contvs}     & 64.70  & 55.67  \\
SPADE~\cite{spade} + AF~\cite{appflow}     & 63.28   & 54.05   \\ \midrule
Target GT Images             & 77.47  & 69.67    \\ 
\bottomrule    
\end{tabular}%
\caption{\textbf{Semantic evaluations on CARLA~\cite{dosovitskiy2017carla} dataset.} Metrics are computed using a single randomly chosen style frame per  sequence. This test shows that our method handles arbitrary style images. Our Class Accuracy and Mean IoU results here are also close the results achieved using the source-view color image as style input. As reported in the main paper paper, using source-view color image as style we obtain $\text{Cls. Acc. of } 77.34 \text{ and mean IoU of } 66.43$.}
\label{table:carla_style}
\end{table}

\section{Network Architectures}
\label{sec:architectures}
\vspace{-3mm}
\noindent \textbf{Semantic Uplifting Network.}
Our Semantic Uplifting Network~(SUN), is a 2D encoder-decoder network with 3 outputs: lifted semantics, MPI transparency $\alpha$ and association function $\Phi$. These 3 outputs are predicted jointly, with the network architecture shown in Table~\ref{table:sun}. 

\noindent \textbf{Appearance Decoder Network.}
Table~\ref{table:adn} shows the detailed architecture of the Appearance Decoder Network~(ADN) network.

\noindent \textbf{Layered Translation Network.}
Layered Translation Network~(LTN) consists of layered appearance generator and style encoder networks. The architecture of both the appearance generator and encoder networks are similar to those used in the SPADE~\cite{spade} paper. However, in this work the generator network performs layered translation of $k$ semantic maps. The output of the network is also different since we predict higher dimensional features, not color images. Our generator, takes input semantics of shape $[H\times W \times k \times l]$ and produces appearance features with dimensions $[H \times W\times l \times f]$. In our experiments, we found that $k=3$ and $f=20$ suffice to achieve good results. The GAN part of our work is based on SPADE, therefore, we use the same discriminator network and training losses as in the SPADE work.

\section{Additional Experimental Details}
\label{sec:training}
\vspace{-3mm}

\subsection{Dataset Details}
\vspace{-2mm}
We use three publicly available datasets for our GVS experiments. In all 3 datasets we used images down scaled to a resolution of $256\times256$ pixels.

\textbf{CARLA.} Using the CARLA~\cite{dosovitskiy2017carla} open source simulation environment, we captured 20 independent sequences in 5 towns. We use 16 sequences for training and 4 sequences for testing, taking one test sequence per city. We capture data using camera arrays mounted on top of a car. Each camera captures color, semantic and depth images. We use 3 groups of camera arrays: \textit{horizontal}, \textit{forward} and \textit{side}-camera groups. The horizontal camera array has 5 cameras at uniform spacing along the $x-axis$. The forward camera array has 5 cameras uniformly distributed along $z-axis$. The side camera array contains 3 horizontally shifted cameras facing the side-view of the car. The spacing between cameras within each array is $54 cms$. During training and testing we take a random pair of cameras from one of the camera groups and use one as source and the other as target.

\textbf{Cityscapes} is a publicly available dataset of urban scenes captured across several German cities~\cite{cordts2016cityscapes}. We use 2975 scenes for training and 500 scenes for test. Each scene is captured with a stereo pair. During training and testing phases, we use one of the two stereo cameras as source and the other as target. Ground truth semantics is available for the left camera images only. We label the right camera images using a pre-trained semantic segmentation network ~\cite{semantic_cvpr19}. Citycapes dataset has no ground-truth depth. We generate depth maps by training the DPS~\cite{im2019dpsnet} network in a self-supervised manner. In order to achieve results which are comparable with SPADE~\cite{spade}, we use instance masks in our experiments on this dataset. Since right camera instance masks are not available, we generate a pseudo ground-truth instance masks by warping the left image masks using the depths estimated by DPS network. While warping, we perform forward-backward depth tests to detect occlusions and in areas where occlusion is detected we leave the instance masks empty. In all of our experiments, we use 19 class labels provided by the dataset.

\textbf{Virtual-KITTI-2} is a synthetic dataset of urban scenes captured with a stereo camera. Each stereo pair has color images together with ground-truth depths and semantic segmentations. The dataset has 6 sequences captured under 5 weather conditions. Each sequence is randomly divided in to train and test sub-sequences. The training sub-sequence covers 80 \% of the frames and the test sub-sequence covers has 20\% of the frames.

\subsection{Training Details}
\vspace{-2mm}

\noindent \textbf{Depth Loss.}
We compute the depth reconstruction loss $\mathcal{L}_{dep}$ on a scaled version of the predicted and target inverse depth maps. The scaled inverse depth is computed as $f_{x} \times 0.54 / depth$, where $f_x$ is the focal length. This is equivalent to converting \textit{depth} into \textit{disparity} by assuming imaginary stereo camera with a baseline of $54 cms$. Since, all of the datasets used in this paper are large scale outdoor scenes, the same scaling works well for all the datasets.

\noindent \textbf{Loss Weighting.}
We set the depth loss weighting factor $\lambda_{1}=0.1$, while the other weighting factors~($\lambda_{0}, \lambda_{2} \text{ and } \lambda_{3}$) are all set to 1. 

\noindent \textbf{Training Protocol}
GVSNet is trained in two phases. In the first phase, the Semantic Uplifting Network~(SUN) is trained using depth reconstruction and semantic alignment losses. For the CARLA~\cite{dosovitskiy2017carla} dataset, we train the SUN network for $30$ epochs. In Cityscapes~\cite{cordts2016cityscapes} dataset, we train for $200$ epochs and in Virtual KITTI-2~\cite{cabon2020virtual}, we train for 60 epochs. In all datasets, we use mini-batch size 12 and Adam~\cite{kingma2014adam} optimizer with a $lr=0.0004$, $\beta_{1}=0.900$, $ \beta_{2}=0.999)$, and $eps=e{-08}$. This SUN training is performed using 3 NVIDIA GTX-2080-Ti GPUs.

In the second phase, the Layered Translation Network (LTN) and Appearance Decoder Network (ADN), are trained by minimising the color and GAN losses. We train these networks while keeping the SUN network fixed. For CARLA dataset, we train for $20$ epochs. In Cityscapes and Virtual KITTI-2 datasets, we train for  $250$ and $35$ epochs respectively. In this phase, we use a batch size of $16$. The training is done using 8 NVIDIA GTX-2080-Ti GPUs. We use Adam~\cite{kingma2014adam} optimizer with the initial learning rate of $lr=0.0004$. The learning rate is kept fixed for the first half of the training. In the second half, we start decreasing the learning rate linearly so that it reaches $0$ at the last iteration.

\noindent \textbf{Instance Masks}
In order to make our Cityscapes~\cite{cordts2016cityscapes} results comparable to SPADE~\cite{spade}, we use instance masks in addition to semantic maps. The instance masks are fed to GVSNet in the following fashion. We compute a one channel instance mask image, as a gradient of the original instance segmentation. This single channel image is concatenated to the lifted/layered semantics output of the Semantic Uplifting Network and given to the Layered Translation Network. Note, that in other datasets we do not use instance masks.

\begin{table}[htb!]
\centering
\small
\begin{tabular}{cccc}
\hline
Layer        & input                & in\_chans & out\_chans \\ \midrule
conv\_0      & appearance features  & f         & 16         \\ \midrule
conv\_1      & conv\_0                                                        & 16        & 32         \\ \midrule
conv\_2      & conv\_1                                                        & 32        & 64         \\ \midrule
conv\_3      & conv\_2                                                        & 64        & 64         \\ \midrule
conv\_4      & conv\_3                                                        & 64        & 64         \\ \midrule
d\_conv\_4   & conv\_4                                                        & 64        & 64         \\ \midrule
d\_conv\_3   & d\_conv\_4 $\bigoplus$ conv\_3                                           & 128       & 32         \\ \midrule
d\_conv\_2   & d\_conv\_3 $\bigoplus$ conv\_2                                           & 64        & 32         \\ \midrule
d\_conv\_1   & d\_conv\_2 $\bigoplus$ conv\_1                                           & 64        & 16         \\ \midrule
output\_conv & d\_conv\_1 $\bigoplus$ conv\_0                                           & 32        & 3          \\ \bottomrule
\end{tabular}%
\caption{Architecture of the Appearance Decoder Network. In this table in\_chans and out\_chans stand for the number of input and output channels. The $\bigoplus$ sign indicates channel-wise concatenation.  The ADN network gets $f-channel$ appearance feature as input and returns a $3-channel$ color image as output. Every layer in this network is a convolutional layer with $3\times\,3$ kernel and \textit{stride} of 1. All layers except output\_conv  have spectral normalisation and LeakyReLU non-linearity~(with negative slope value of $-0.2$) after convolution. output\_conv layer applies Tanh non-linearity after convolution and it has no spectral normalisation~\cite{miyato2018spectral}. Layers conv\_0 up to conv_4 down scale the spatial dimensions of their output using a  bilinear down-sampling kernel. Equivalently, layers d\_conv_4 up to d_conv_1 perform bilinear up-sampling with a factor of 2, before convolution is applied.}
\label{table:adn}
\end{table}

\begin{table}[htb!]
\centering
\resizebox{\textwidth}{!}{%
\scriptsize
\begin{tabular}{cccccc}
\toprule
Layer &
  Input &
  Type &
  Stride &
  in\_chans &
  out\_chans \\ \midrule
conv1a        & input sem       & Conv2d + ReLU    & 2 &  l                        & 32                       \\ \midrule
conv1b        & conv1a          & Conv2d + ReLU    & 1 & 32                       & 32                       \\ \midrule
conv2a        & conv1b          & Conv2d + ReLU    & 2 & 32                       & 64                       \\ \midrule
conv2b        & conv2a          & Conv2d + ReLU    & 1 & 64                       & 64                       \\ \midrule
conv3a        & con2b           & Conv2d + ReLU    & 2 & 64                       & 128                      \\ \midrule
conv3b        & conv3a          & Conv2d + ReLU    & 1 & 128                      & 128                      \\ \midrule
conv4a        & conv3b          & Conv2d + ReLU    & 2 & 128                      & 256                      \\ \midrule
conv4b        & conv4a          & Conv2d + ReLU    & 1 & 256                      & 256                      \\ \midrule
conv5a        & conv4b          & Conv2d + ReLU    & 2 & 256                      & 512                      \\ \midrule
conv5b        & conv5a          & Conv2d + ReLU    & 1 & 512                      & 512                      \\ \midrule
conv6a        & conv5b          & Conv2d + ReLU    & 2 & 512                      & 512                      \\ \midrule
conv6b        & conv6a          & Conv2d + ReLU    & 1 & 512                      & 512                      \\ \midrule
conv7a        & conv6b          & Conv2d + ReLU    & 2 & 512                      & 512                      \\ \midrule
conv7b        & conv7a          & Conv2d + ReLU    & 1 & 512                      & 512                      \\ \midrule
dconv7        & conv7a          & Conv2d + ReLU    & 1 & 512                      & 512                      \\ \midrule
dconv6        & dconv7 $\bigoplus$ conv6b & Conv2d + ReLU    & 1 & 1024                     & 512                      \\ \midrule
dconv5        & dconv6 $\bigoplus$ conv5b & Conv2d + ReLU    & 1 & 1024                     & 512                      \\ \midrule
dconv4        & dconv5 $\bigoplus$ conv4b & Conv2d + ReLU    & 1 & 768                      & 384                      \\ \midrule
dconv3        & dconv4 $\bigoplus$ conv3b & Conv2d + ReLU    & 1 & 512                      & 256                      \\ \midrule
dconv2        & dconv3 $\bigoplus$ conv24 & Conv2d + ReLU    & 1 & 320                      & 96                       \\ \midrule
dconv1        & dconv2 $\bigoplus$ conv1b & Conv2d + ReLU    & 1 & 128                      & 96                       \\ \midrule
base\_1       & dconv1          & Conv2d + ReLU    & 1 & 96                       & 96                       \\ \midrule
base\_2       & base\_1         & ResBlock         & 1 & 96                       & 96                       \\ \midrule
base\_3       & base\_2         & ResBlock         & 1 & 96                       & 96                       \\ \midrule
base\_4       & base\_3         & ResBlock         & 1 & 96                       & 96                       \\ \midrule
base\_5       & base\_4         & ResBlock with BN & 1 & 96                       & 96                       \\ \midrule
base\_6       & base\_5         & ResBlock with BN & 1 & 96                       & 96                       \\ \midrule
out\_conv\_1a & base\_6         & Conv2d+ReLU        & 1 & 96                       & (l x  (k-1) + m*(k+1))/2 \\ \midrule
out\_conv\_1b & out\_conv\_1a   & Conv2d+ReLU + BN   & 1 & (l x  (k-1) + m*(k+1))/2 & (l x  (k-1) + m*(k+1))/2 \\ \midrule
out\_conv\_a  & out\_conv\_1b   & Conv2d+ReLU        & 1 & (l x  (k-1) + m*(k+1))/2 & (l x  (k-1) + m*(k+1))/2 \\ \midrule
out\_conv\_1b & out\_conv\_a    & Conv2d             & 1 & (l x  (k-1) + m*(k+1))/2 & l x  (k-1) + m*(k+1)     \\ \bottomrule
\end{tabular}%
}
\caption{Semantic Uplifting Network Architecture. In this table \textit{in_chans} and \textit{out_chans} refer to input and output number of channels, respectively. The size of the SUN network depends on the following hyper-parameters: number of layered semantic maps~(k), number of MPI planes~(m) and number semantic classes~(l). The output from the \textit{out\_conv\_1b} layer is split into $\alpha$, $\Phi$ and layered semantics channels. $\alpha$ and $\Phi$ take up $m$ and $k\times m$ channels, respectively. Since we create the layered semantic representation includes the input semantics, the network predicts layered semantics only for $(k-1)$ layers, with a total of $ (k-1) \times l$ channels. Sigmoid activation is applied on the $\alpha$ and $\Phi$ outputs. All of the layers in this network are 2D convolutional layers. Encoder layers $conv1a$ up to $conv7b$ decrease the spatial resolution of their output by a factor of 2 using stride 2 convolutions. Decoder layers $dconv7$ to $deconv1$ apply nearest neighbour up-sampling with a factor of 2 before applying convolution and ReLU. The ResBlock have 2 convolutional layers~(the first one has ReLU activation), and the output of the second layer is added to the input. The $\bigoplus$ sign refers to channel-wise concatenation.}
\label{table:sun}
\end{table}

\end{document}